%% file: main.tex
\documentclass{article}
\usepackage{arxiv}

\usepackage{amsmath}
\usepackage{bm, amsmath, physics, amssymb}

\usepackage[utf8]{inputenc} 
\usepackage[T1]{fontenc}    
\usepackage{hyperref}       
\usepackage{url}            
\usepackage{booktabs}       
\usepackage{amsfonts}       
\usepackage{nicefrac}       
\usepackage{microtype}      
\usepackage{lipsum}		
\usepackage{graphicx}
\usepackage{natbib}
\usepackage{doi}
\usepackage{caption}
\usepackage{algorithmic}
\usepackage[ruled,vlined,linesnumbered]{algorithm2e}
\usepackage{subcaption}
\usepackage{multicol}
\usepackage{multirow}

\title{Sparse and Transferable Universal Singular Vectors Attack}

\author{%
Kseniia Kuvshinova$^{1,4}$\thanks{These authors contributed equally to this work.} \ \thanks{Corresponding author.} \quad Olga Tsymboi$^{1,2}$\footnotemark[1] \quad Ivan Oseledets $^{3, 4}$
\\
$^1$Sber AI Lab, Moscow, Russia \\ $^2$Moscow Institute of Physics and Technology, Moscow, Russia
\\ $^3$Artificial Intelligence Research Institute (AIRI), Moscow, Russia
\\$^4$Skolkovo Institute of Science and Technology, Moscow, Russia\\
\texttt{kse.kuvshinova@gmail.com}, \texttt{tsimboy.oa@phystech.edu}, \texttt{oseledets@airi.net}}

\date{}

\hypersetup{
pdftitle={A template for the arxiv style},
pdfsubject={q-bio.NC, q-bio.QM},
pdfauthor={David S.~Hippocampus, Elias D.~Striatum},
pdfkeywords={First keyword, Second keyword, More},
}

\begin{document}
\maketitle
\input{sec/0_abstract}    
\input{sec/1_intro}
\input{sec/2_framework}
\input{sec/3_exp.tex}
\input{sec/4_related}
\input{sec/5_limitations}
\input{sec/6_conclusions}


\bibliographystyle{unsrtnat}
\bibliography{main}

\appendix
\input{sec/X_suppl}

\begin{abstract}
	\lipsum[1]
\end{abstract}

\keywords{First keyword \and Second keyword \and More}

\section{Introduction}
\lipsum[2]
\lipsum[3]

\section{Headings: first level}
\label{sec:headings}

\lipsum[4] See Section \ref{sec:headings}.

\subsection{Headings: second level}
\lipsum[5]
\begin{equation}
	\xi _{ij}(t)=P(x_{t}=i,x_{t+1}=j|y,v,w;\theta)= {\frac {\alpha _{i}(t)a^{w_t}_{ij}\beta _{j}(t+1)b^{v_{t+1}}_{j}(y_{t+1})}{\sum _{i=1}^{N} \sum _{j=1}^{N} \alpha _{i}(t)a^{w_t}_{ij}\beta _{j}(t+1)b^{v_{t+1}}_{j}(y_{t+1})}}
\end{equation}

\subsubsection{Headings: third level}
\lipsum[6]

\paragraph{Paragraph}
\lipsum[7]

\section{Examples of citations, figures, tables, references}
\label{sec:others}

\subsection{Citations}
Citations use \verb+natbib+. The documentation may be found at
\begin{center}
	\url{http://mirrors.ctan.org/macros/latex/contrib/natbib/natnotes.pdf}
\end{center}

Here is an example usage of the two main commands (\verb+citet+ and \verb+citep+): Some people thought a thing \citep{kour2014real, hadash2018estimate} but other people thought something else \citep{kour2014fast}. Many people have speculated that if we knew exactly why \citet{kour2014fast} thought this\dots

\subsection{Figures}
\lipsum[10]
See Figure \ref{fig:fig1}. Here is how you add footnotes. \footnote{Sample of the first footnote.}
\lipsum[11]

\begin{figure}
	\centering
	\fbox{\rule[-.5cm]{4cm}{4cm} \rule[-.5cm]{4cm}{0cm}}
	\caption{Sample figure caption.}
	\label{fig:fig1}
\end{figure}

\subsection{Tables}
See awesome Table~\ref{tab:table}.

The documentation for \verb+booktabs+ (`Publication quality tables in LaTeX') is available from:
\begin{center}
	\url{https://www.ctan.org/pkg/booktabs}
\end{center}

\begin{table}
	\caption{Sample table title}
	\centering
	\begin{tabular}{lll}
		\toprule
		\multicolumn{2}{c}{Part}                   \\
		\cmidrule(r){1-2}
		Name     & Description     & Size ($\mu$m) \\
		\midrule
		Dendrite & Input terminal  & $\sim$100     \\
		Axon     & Output terminal & $\sim$10      \\
		Soma     & Cell body       & up to $10^6$  \\
		\bottomrule
	\end{tabular}
	\label{tab:table}
\end{table}

\subsection{Lists}
\begin{itemize}
	\item Lorem ipsum dolor sit amet
	\item consectetur adipiscing elit.
	\item Aliquam dignissim blandit est, in dictum tortor gravida eget. In ac rutrum magna.
\end{itemize}

\bibliographystyle{unsrtnat}
\bibliography{references}  






\end{document}

%% file: sec/0_abstract.tex
\begin{abstract}
The research in the field of adversarial attacks and models' vulnerability is one of the fundamental directions in modern machine learning. Recent studies reveal the vulnerability phenomenon, and understanding the mechanisms behind this is essential for improving neural network characteristics and interpretability. 
In this paper, we propose a novel sparse universal white-box adversarial attack. 
Our approach is based on truncated power iteration providing sparsity to $(p,q)$-singular vectors of the hidden layers of Jacobian matrices. 
Using the ImageNet benchmark validation subset, we analyze the proposed method in various settings, achieving results comparable to dense baselines with more than a 50\% fooling rate while damaging only 5\% of pixels and utilizing 256 samples for perturbation fitting.
We also show that our algorithm admits higher attack magnitude without affecting the human ability to solve the task.
Furthermore, we investigate that the constructed perturbations are highly transferable among different models without significantly decreasing the fooling rate. 
Our findings demonstrate the vulnerability of state-of-the-art models to sparse attacks and highlight the importance of developing robust machine learning systems.
\end{abstract}

%% file: sec/1_intro.tex
\section{Introduction}
\label{sec:intro}

In recent years, deep learning approaches have become increasingly popular in many areas and applications, starting from computer vision \cite{dosovitskiy2021image} 
and natural language processing \cite{touvron2023llama, chung2022scaling} to robotics \cite{roy2021machine} and speech recognition \cite{baevski2020wav2vec}. 
The success and availability of pre-trained neural networks have also made it easier for researchers and developers to use these models for their applications.
Despite tremendous advances, it was discovered that deep learning models are vulnerable to small perturbations of input data called adversarial attacks that mislead models and cause incorrect predictions \cite{szegedy2013intriguing, Goodfellow2014, Moosavi-Dezfooli_2017_CVPR}. 
Adversarial attacks as a phenomenon first appeared in the field of computer vision 
and have raised concerns about the reliability in safety-critical machine learning applications. 

Initially, adversarial examples were constructed for each individual input \cite{szegedy2013intriguing}, making it challenging to scale attacking methods to large datasets.
In \cite{Moosavi-Dezfooli_2017_CVPR}, the authors show the existence of universal adversarial perturbations (UAPs) that result in the model's misclassification for most of the inputs. Such attacks are crucial for adversarial machine learning research, as they are easier to deploy in real-world applications and raise questions about the safety and robustness of state-of-the-art architectures. 
However, the proposed optimization algorithm requires vast data, making it complicated to fool real-world systems. In contrast, \cite{Khrulkov_2018_CVPR} proposes a sample-efficient method to construct perturbation using leading $(p, q)$-singular vectors \cite{boyd1974power} of the Jacobian of the hidden layer. However, Jacobian is infeasible directly due to the memory limitation, the authors overcomes this issue using the generalized power method for the attack computation. 


Above approaches formalize imperceptibility using straightforward vector norm constraints in the underlying optimization problem. 
However, in general, an attack can alter the image significantly, leaving its semantics unchanged \cite{NEURIPS2018_8cea559c, brown2018unrestricted}. One can step beyond the small-norm imperceptibility definition and perform a patch attack as a physical sticker on an object in real-time conditions \cite{hu2022adversarial, li2019adversarial, pautov2019adversarial, kaziakhmedov2019real}. In this work, we focus on attacks under sparsity constraints. Indeed, damaging a small number of pixels does not significantly influence 
image semantics and hence
human 
perception, resulting in unchanged prediction.



There are quite a lot methods to compute sparse adversaries \cite{DBLP:journals/corr/abs-1909-05040, modas2019sparsefool, yuan2021meta, dong2020greedyfool}, however, 
the transferability of $l_0$-~bounded 
attacks is still low \cite{papernot2016transferability, he2022transferable, zhang2023improving}. 
In other words, these methods may perform poorly in grey-box settings (when, instead of the initial model, a surrogate model is attacked).
However, we should highlight that only a few works aim to 
incorporate sparsity constraints into universal attack setup \cite{shafahi2020universal, Croce_Andriushchenko_Singh_Flammarion_Hein_2022}
. More than that, an auxiliary generative model is usually used to construct such transferable sparse attacks  \cite{he2022transferable, hayes2018learning, mopuri2018nag}. 



The main focus of this paper is to investigate computer vision 
models' robustness to sparse universal adversarial examples. Summing up, our main contribution is as follows:
\begin{itemize}
    \item We propose a new approach to construct sparse UAPs on hidden layers subject to predefined cardinality or sparsity patterns. 
    
    \item We assess our method on the ImageNet benchmark dataset \cite{5206848} and evaluate it on various deep learning models. We compare it against existing universal approaches regarding the fooling rate and the transferability between models. 
    
    \item In our experimental study, we show that the proposed method produces highly transferable perturbations. It is important that our approach is efficient with respect to sample size~-- a moderate sample size of 256 images to construct an attack on is enough for a reasonable attack fooling rate.
\end{itemize}

%% file: sec/2_framework.tex
\section{Framework}
\label{sec:formatting}

In this paper, we focus on the problem of untargeted universal perturbations 
for image classification.
The problem of universal adversarial attacks can be framed as finding a perturbation $\varepsilon$ that, when added to most input images $x$, causes the classifier to predict a different class than it would for the original images. Let $f: \mathcal{X} \to \mathcal{Y}$ be a classification model defined for the dataset $\mathcal{D} = \{x_i, y_i\}_{i = 1}^N$, where $x_i$ is an input and $y_i$ is a corresponding label, then, according to \cite{7780651}, UAP is a perturbation $\varepsilon$ such that
\begin{equation*}
    \mathbb{P}_{x \sim \mu}[f(x + \varepsilon) \neq f(x)] \geq 1 - \delta, \quad \|\varepsilon\| \leq \xi,
\end{equation*}
where $\mu$ denotes a distribution of input data, $1 - \delta$ is the minimal Fooling Rate (FR) and $\xi$ is the attack magnitude. It should be highlighted that this perturbation must not change human prediction, meaning that the true class of the attacked image remains unchanged, but a small norm constraint could be omitted \cite{NEURIPS2018_8cea559c, brown2018unrestricted}. 

Adversarial perturbations are often obtained via optimization problem solution. 
The most straightforward approach is to maximize expected cross-entropy loss $\mathcal{L}(x + \varepsilon, y)$.
It was shown \cite{Khrulkov_2018_CVPR} that instead of attacking the model output, one can attack hidden layers of a model. Then the error produced in this way propagates to the last network layer, resulting in a model prediction change. 
Given the $i$-th layer $l$, the optimization problem, in this case, can be obtained via Taylor expansion:
\begin{gather}
    l(x + \varepsilon) - l(x) \approx J_i(x)\varepsilon\nonumber\\
    \|l(x + \varepsilon) - l(x)\|_q^q \to \max_{\|\varepsilon\|_p \leq \xi},\label{eq:layer-wise-uap}
\end{gather}
where $J_i(x)$ is the $i$-th layer Jacobian operator and $J_i(x)\varepsilon$ is Jacobian action on $\varepsilon$. Finally, \eqref{eq:layer-wise-uap} is equivalent to the following problem.
\begin{gather}\label{eq:lw-uap-opt}
 \max_{\varepsilon}\mathbb{E}_{x \sim \mu}\|J_i(x)\varepsilon\|^q_q, \quad s.t.~\|\varepsilon\|_p = 1.
\end{gather}
The solution to \eqref{eq:lw-uap-opt} can be referred to as Jacobian $(p, q)$-singular vector and defined up to the signed scale factor, here $p$, $q$ are the hyperparameters to be tuned, and expectation is relaxed via averaging over a batch.

\textbf{Our approach.}
In this paper, we incorporate the universal layerwise approach from above with sparsity or, formally speaking, additional non-convex $l_0$ constraint:
\begin{equation}
    \begin{split}
        \sum_{x \in \text{batch}}\abs{\abs{J_i(x) \varepsilon}}_q^q \to \max, \\
        s.t.~\abs{\abs{\varepsilon}}_p  = 1, \quad\abs{\abs{\varepsilon}}_0 \le k.
    \end{split}
    \label{eq:sparse_problem}
\end{equation}
In the case when $p = q = 2$, the problem above leads to the famous problem of finding sparse eigenvalues. 
However,  for an arbitrary pair $(p,q)$, it is a non-convex and NP-hard problem. 
One way to obtain an approximate solution is to use the truncated power iteration method (TPower, \cite{https://doi.org/10.48550/arxiv.1112.2679}).
This method can effectively recover the 
sparse eigenvector for symmetric positive semidefinite matrices and when the underlying matrix admits sparse eigenvectors. Despite efficiency and simplicity, the major TPower drawback is the theoretical guarantee with a narrow convergency region. One way to reduce the effect of this issue is to reduce the number of nonzero entries iteratively. 

In this paper, we introduce an algorithm that adopts TPower for the case of arbitrary $p, q$ and effectively solves the problem of universal perturbation finding. 

Let us rewrite \eqref{eq:sparse_problem} using the dual norm definition, then we obtain
\begin{equation}
    \begin{split}
        \max_{\varepsilon \in \mathcal{B}_p(1)} 
        \max_{y \in \mathcal{B}_{q^*}(1)} 
        y^{\top} J_i \varepsilon, 
        \quad s.t.~\abs{\abs{\varepsilon}}_0  \le k,
    \end{split}
    \label{eq:sparse_dual}
\end{equation}
where ${(q^*)}^{-1} + q^{-1} = 1$, $\mathcal{B}_p(1) = \{x \in \mathbb{R}^n|~\|x\|_p = 1\}$ and $J_i = [J_i(x_1))^{\top}, \ldots, J_i(x_N))^{\top}]^{\top},~x_j \in \text{batch}$. The solution could be found via Alternating Maximization (AM) Method.

For any fixed perturbation vector $\varepsilon$, the inner problem is linear and admits a closed-form solution, using $b = J_i\varepsilon$ for notation, we have:
\begin{equation}
    y = \frac{\psi_q(b)}{\|\psi_q(b)\|_{q^*}}, \quad \psi_q(y) = \mathrm{sign} (y) \abs{y}^{q - 1}.
\label{eq:dual_solution}
\end{equation}

Changing the order of maximizations in \eqref{eq:sparse_dual}, the subproblem for $\varepsilon$ remains the same except $l_0$ constraint, which could be replaced by additional binary variable $t$ maximization. Thus, denoting $d = J_i^{\top} y$, we have:
\begin{gather}
    \max_{\varepsilon \in \mathcal{B}_p(1)} \; d^{\top} \varepsilon, \quad s.t.~\abs{\abs{\varepsilon}}_0 \le k.
\end{gather}
\begin{gather}
    \max_t\max_{\varepsilon \in \mathcal{B}_p(1)} \; (t \cdot d)^{\top} \varepsilon, \quad s.t.~t_j \in \{0, 1\},~\forall j.
    \label{eq:binary_var_problem}
\end{gather}
here $(\cdot)$ is a Hadamar product. 

For a fixed $t$, the problem is reduced to the previous case, and hence
\begin{equation}
\varepsilon \sim \psi_{p^*}(t \cdot d),
\end{equation}
Now, substituting this equation  to the objective \eqref{eq:binary_var_problem}, we derive:
\begin{align*}
    (t \cdot d)^{\top}{\varepsilon} 
    &= d^{\top}{\rm diag}(t)|d|^{p^* - 1}{\rm sign}(d) = \sum_{j}|d_j||d_j|^{p^* - 1}t_j = \sum_j |d_j|^{p^*}t_j \to \max_t
\end{align*}
and thus maximization by $t$ is simply a selection of the greatest components of vector $d$ in absolute value, which could be done by truncation operator:
\begin{equation}
    T_{p^*, k} (d) = 
    \begin{cases}
        d_i, \quad i \in \mathrm{ArgTop}_k \{\abs{\abs{d_i}}_{p^*}\}, \\
        0, \quad \text{otherwise},
    \end{cases}
\end{equation}
here $\|\cdot\|_{p^*}$ is used to emphasize the possibility of a block-sparse solution under the predefined sparsity pattern for patch attack.

Finally, putting it together, we derive the following alternating maximization update at the step $s$ for attack training.:
\begin{equation}
    \varepsilon^{s + 1} = T_{p^*, k} \left[~\sum_{x \in \text{batch}} J_i^{\top}(x) \psi_q (J_i(x) \varepsilon^s) \right],
    \label{eq:iteration}
\end{equation}
\begin{equation}
    \varepsilon^{s + 1} = \frac{\psi_{p^*}(\varepsilon^{s + 1})}{\abs{\abs{\psi_{p^*}(\varepsilon^{s + 1})}}_p}.
    \label{eq:iteration_norm}
\end{equation}
The overall algorithm is presented in Algorithm~\ref{alg:TPowerAttack}, where we gradually decrease the cardinality through the iterations to enhance convergency.
\begin{algorithm}
    \caption{TPower Attack}
    \label{alg:TPowerAttack}
    \begin{algorithmic}[1]
    \REQUIRE $n\_steps$, $init\_truncation \in (0, 1)$, batch of images $x_j \in \mathbb{R}^{n \times n \times \text{channels}}$, $j \in \overline{1, N}$, $q$, $p$, target cardinality $top\_k$, $patch\_size$, $reduction\_steps$. 
    \STATE $k = init\_truncation \cdot (n // patch\_size)^2 \cdot \text{channels}$. 
    \STATE $k = \max(k, top\_k)$.
    \STATE $\varepsilon =$ random tensor of batch size.
    \FOR{$s$ from 1 to $n\_steps$}
        \STATE $\varepsilon^{s + 1} = T_{p^*, k} \left[~\sum\limits_{x \in \text{batch}} J_i^{\top}(x) \psi_q (J_i(x) \varepsilon^s) \right], \quad \eqref{eq:iteration}$
        \STATE $\varepsilon^{s + 1} = \frac{\psi_{p^*}(\varepsilon^{s + 1})}{\abs{\abs{\psi_{p^*}(\varepsilon^{s + 1})}}_p}, \quad \eqref{eq:iteration_norm}$
        \IF{$s\mod{reduction\_steps} = 0$}
            \STATE $k_{\text{reduction}} = 
    \mathrm{pow}(\frac{k}{top\_k},{\frac{reduction\_steps}{n\_steps}})$
            \STATE $k = \max(k/k_{\text{reduction}},~ top\_k)$
        \ENDIF
    \ENDFOR
    \ENSURE $\varepsilon$
    \end{algorithmic}
\end{algorithm}

%% file: sec/3_exp.tex
\section{Experiments}

This section presents the experiments conducted to analyze the effectiveness of sparse UAPs described above. The experiments were implemented using PyTorch, and the code will be made publicly available on Github upon publication.

\subsection{Experiments setup}
\textbf{Datasets.} In this work, following \cite{Khrulkov_2018_CVPR}, to evaluate the performance of the proposed sparse attack, we used the validation subset of the ImageNet benchmark dataset (ILSVRC2012, \cite{deng2009imagenet}), which contains 50,000 images belonging to 1,000 object categories.
We randomly sample 256 images from the ImageNet validation subset for attack training. 
Additionally, to perform grid search, there were stratified sampled 5000 images as a validation set, while the rest of the samples were used as the test set.

\textbf{Models.}
During the empirical analyses, we restrict ourselves to the following models to be examined: DenseNet161 \cite{huang2017densely}, EffecientNetB0, EffecientNetB3 \cite{tan2019efficientnet}, InceptionV3 \cite{szegedy2015going}, ResNet101, ResNet152 \cite{he2016deep}, VGG19 \cite{DBLP:journals/corr/SimonyanZ14a},  Wide ResNet101 \cite{DBLP:conf/bmvc/ZagoruykoK16}, DEIT~base \cite{touvron2021training}, ViT~base \cite{dosovitskiy2020vit}. 
For each model, there were used ImageNet pre-trained checkpoints. 

\textbf{Hyperparameters.}
In our experiments, to estimate attack performance, we vary the following hyperparameters: model, layer to be attacked, patch size $\in \{1, 4, 8\}$ and objective norm parameter $q \in \{1, 2, 3, 5, 7, 10\}$ while keeping $p$ fixed, in particular, $p = \infty$ which is motivated by the previous study \cite{Khrulkov_2018_CVPR}. 
The number of non-zero patches $k$ is also fixed in accordance with the image and patch sizes and selected so that the fraction of damaged pixels is equal to 5\%, which further allows us to increase the attack magnitude up to 1 (Table~\ref{table:1}). 
We gradually went through all semantic blocks to study the performance dependence on the layer to be attacked (see Appendix for more details).

We performed the computation on four GPU's NVIDIA A100 of 80GB. The full grid search for TPower Attack took 765 GPU hours for nine models in total (6 values for $q$, ten values for $\alpha$ and three patch sizes).

\textbf{Evaluation metrics.}
For evaluation, we report Fooling Rate (FR) \eqref{eq:FR} on the validation and test subsets for the best perturbation obtained on the 256 training samples.
It also means that we find ourselves in an unsupervised setting and do not need access to the ground truth labels.
\begin{equation}
    FR = \frac{1}{N}\sum\limits_{x \in \mathcal{D}}[f(x) \neq f(x + \varepsilon)]
    \label{eq:FR}
\end{equation}
Along with fooling rate, we focus our consideration on Attack Success Rate (ASR), namely the portion of misclassified samples after the attack performance filtered subject to the initial model's correct predictions:
\begin{equation}
    ASR = \frac{\sum\limits_{x \in \mathcal{D}}[f(x) \neq f(x + \varepsilon)][f(x) = y]}{\sum\limits_{x, y \in \mathcal{D}}[f(x) = y]}
    \label{eq:ASR}
\end{equation}


\subsection{Main results}
We train our attack on nine different models and compare it to the stochastic gradient descent (SGD) attack~\cite{shafahi2020universal} and the dense analogue of our approach proposed by~\cite{Khrulkov_2018_CVPR}, here and below, we refer the last approach as singular vectors (SV) attack. We also consider transferability setup and discover the FR dependence on $q$. 
Following previous research, which relies on the small norm assumption, the magnitude was decreased to $10/255$ for dense baselines. 
Our approach is significantly superior to dense attacks. Poor results in the SGD can be explained by the fact that a relatively large train set size is required to obtain an efficient attack (e.i. greater than the number of classes), while for our proposed approach, 256 images are enough.

The results of the grid search are presented in Table~\ref{table:1}, where we report optimal hyperparameters for each model with respect to validation FR. 
For this setting, the comparison with the baselines is provided in Table~\ref{table:2}, where for SV attack, we additionally perform a similar grid search on the layer and $q$.

\begin{table*}[h!]
\centering
\begin{tabular}{lcccccc}
 \toprule
 Model & Top k & Patch Size & $q$ & Attacked Layer & Test ASR & Test FR \\
 \midrule
 DenseNet161 & 2509 & 1 & 1 & features.denseblock2.denselayer6 & 87.61 & 89.11 \\
 EffecientNetB0 & 2509 & 1 & 1 & features.2.1.block & 29.66 & 37.09 \\
 EffecientNetB3 & 4500 & 1 & 1 & features.1.0.block & 9.66 & 15.22 \\
 InceptionV3 & 4471 & 1 & 1 & maxpool2 & 82.83 & 85.04 \\
 ResNet101 & 157 & 4 & 1 & layer2.3 & 93.8 & 94.57 \\
 ResNet152 & 157 & 4 & 1 & layer2.3 & 94.24 & 94.84 \\
 WideResNet101 & 157 & 4 & 1 & layer3.1 & 93.71 & 94.36 \\
 DEIT base & 2509 & 1 & 1 & vit.encoder.layer.0 & 36.12 & 43.37 \\
 ViT base & 2509 & 1 & 1 & vit.encoder.layer.0 & 46.76 & 52.5 \\
\bottomrule
\end{tabular}
\caption{Metrics and hyperparameters for the best-performed sparse UAPs for each model.}
\label{table:1}
\end{table*}

Our TPower attack approach outperforms baselines for almost all models except EfficientNets and demonstrates diverse attack patterns.

\begin{figure*}[h!]
    \centering
    \begin{subfigure}
        {0.49\textwidth}
        \includegraphics[width=0.49\textwidth]{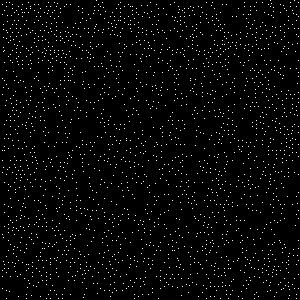}
        \includegraphics[width=0.49\textwidth]{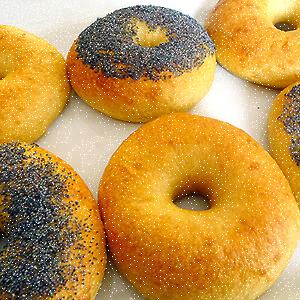}
        \caption{EfficientNetB3}
    \end{subfigure}
    \begin{subfigure}
        {0.49\textwidth}
        \includegraphics[width=0.49\textwidth]{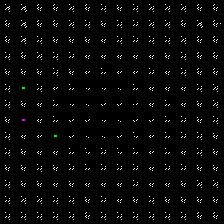}
        \includegraphics[width=0.49\textwidth]{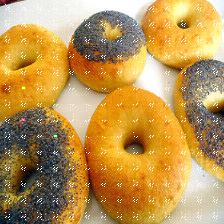}\caption{DEIT Base}  
    \end{subfigure}    
    \caption{Sparse UAPs obtained using the TPower algorithm and corresponding examples of attacked images. 
    Perturbations were computed using the best-performed layers on gridsearch.
    }
    \label{fig:images}
\end{figure*}

\begin{figure*}[h!]
    \centering
    \begin{subfigure}
        {0.32\textwidth}
        \includegraphics[width=\textwidth]{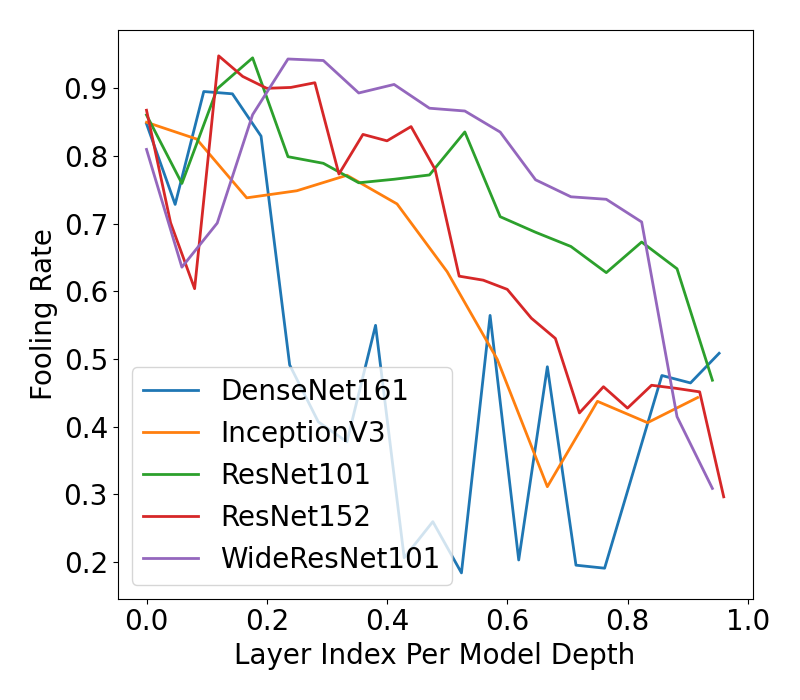}
        \caption{}
        \label{fig:layer_1}
    \end{subfigure}
    \begin{subfigure}
        {0.32\textwidth}
        \includegraphics[width=\textwidth]{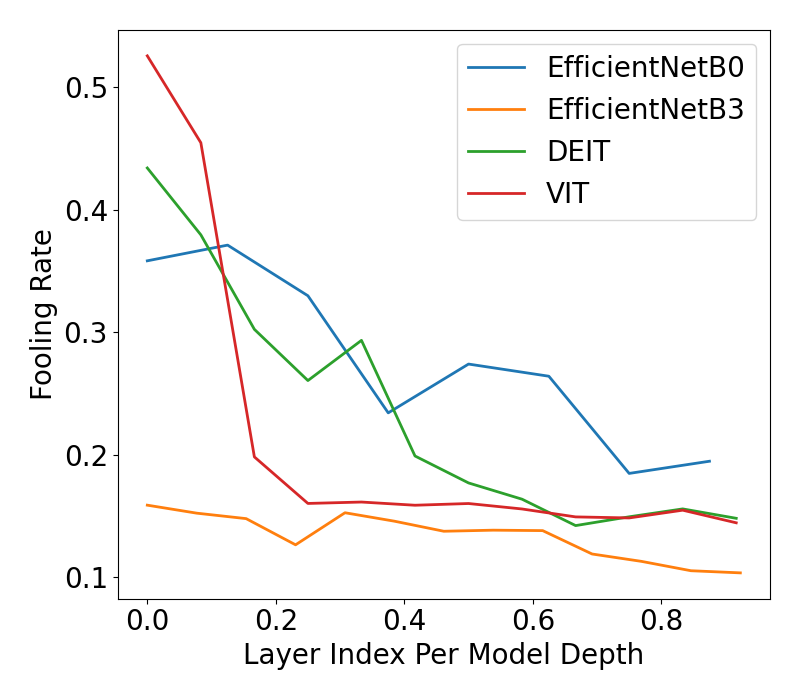}
        \caption{}
        \label{fig:layer_2}
    \end{subfigure}
    \begin{subfigure}
        {0.32\textwidth}
        \includegraphics[width=\textwidth]{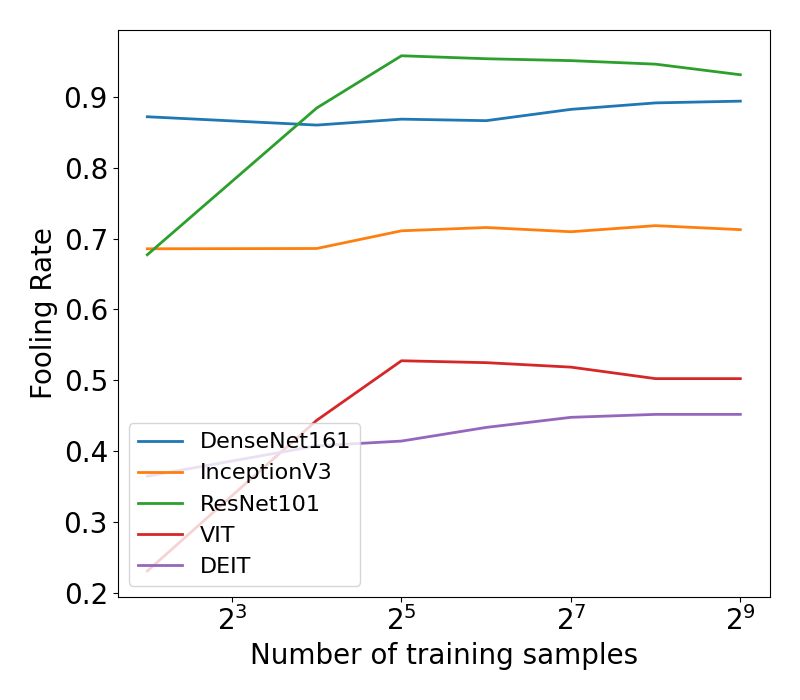}
        \caption{}
        \label{fig:train_size}
    \end{subfigure}
    \caption{\ref{fig:layer_1} and  \ref{fig:layer_2}: 
    Fooling Rate (FR) 
    dependence on layer ratio for examined models. \ref{fig:train_size}: The example of fooling rate saturation depending on training set size for optimal hyperparameters; here, one can observe that 256 is the worst case amount among most vulnerable models.
    }
    \label{fig:layer number}
\end{figure*}

From Table~\ref{table:2}, one can conclude that EfficientNet is the most robust architecture. Some architectural choices, like compound scaling, limit the gradient flow during backpropagation. This fact makes it more challenging for attackers to generate efficient adversarial perturbations.
It is worth mentioning that the attack training on EfficientNets results in noticeably different perturbation patterns than other models. 
It is more similar to white noise (see Figure~\ref{fig:images}), whereas the distribution of the patches trained on other models resembles higher order Sobol points \cite{sobol1967distribution}.
Additionally, for the ViT model, Figure~\ref{fig:images} demonstrates a highly interpretable pattern. 
Formally speaking, during ViT preprocessing, the image is cut into fixed-size patches, which are further flattened and combined with positional encoding \cite{wu2020visual}. 
Indeed, perturbation forms a quasi-regular grid repeating the locations of the patch junctions.
Moreover, attacking lower, more sensitive to preprocessing by construction layers causes the highest model vulnerability (Figure~\ref{fig:layer number}). 

\begin{table}[!htb]
    \centering
    \begin{minipage}[t]{.48\linewidth}\vspace{0pt}
      \centering
\begin{tabular}{lcccc}
 \toprule
 Model & \textbf{Ours} & SV & SGD & LMax \\
 \midrule
 DenseNet161 & \textbf{89.11} & 34.25 & 15.53 & 22.70 \\
 EffecientNetB0 & \textbf{37.09} & 34.44 & 16.35 & - \\
 EffecientNetB3 & \textbf{15.22} & 13.49 & 8.22 & 11.10 \\
 InceptionV3 & \textbf{85.04} & 27.88 & 13.11 & 23.63 \\
 ResNet101 & \textbf{94.57} & 50.05 & 17.17 & 44.51 \\
 ResNet152 & \textbf{94.84} & 35.93 & 15.05 & 20.12 \\
 WideResNet101 & \textbf{94.36} & 36.35 & 15.02 & 26.56 \\
 DEIT base & \textbf{43.37} & 31.1 &  19.16 & 23.61 \\
 ViT base & \textbf{52.5} & 26.01 & 17.09 & 29.54 \\
\bottomrule
\vspace{\fill}
\end{tabular}
\captionof{table}{Comparison between TPower (Ours), SV and SGD adversarial perturbations. For the TPower and SV attack, there reported the 
test Fooling Rate (FR) 
for optimal hyperparameters after the gridsearch.}
\label{table:2}
    \end{minipage}%
    \hfill
    \begin{minipage}[t]{.48\linewidth}\vspace{0pt}
      \centering
\begin{tabular}{lcccccc}
 \toprule
 Model / Patch size & 1 & 4 & 8 \\
 \midrule
 DenseNet161 & \textbf{89.11} & 78.52 & 64.82 \\
 EffecientNetB0 & \textbf{37.09} & 29.95 & 22.87 \\
 EffecientNetB3 & \textbf{15.22} & 13.11 & 13.28 \\
 InceptionV3 & \textbf{85.04} & 22.36 & 77.66 \\
 ResNet101 & 89.85 & \textbf{94.57} & 83.48 \\
 ResNet152 & 89.53 & \textbf{94.84} & 76.78 \\
 WideResNet101 & 93.97 & \textbf{94.36} & 84.1 \\
 DEIT base & \textbf{43.37} & 22.37 & 15.59 \\
 ViT base & \textbf{52.5} & 15.44 & 14.54 \\
\bottomrule
\vspace{\fill}
\end{tabular}
\vfill
\captionof{table}{Tpower attack 
Fooling Rate (FR) 
dependence on patch size on test 
for frozen other optimal hyperparameters.}
\label{app:table:2}
    \end{minipage} 
\end{table}


\textbf{Dependence on patch size.} Our empirical study shows that, in general, lower patch size values are more beneficial in terms of FR (see Table~\ref{table:1}). One can see that pixel-wise attack mode is more efficient regarding the fooling rate. This might be related to the fact that uniform square greed is not an optimal sparsity pattern. However, for most models, the decrease in performance is not dramatic, except for the transformers one. For those models where the optimal patch size option is 4, FR does not decrease significantly compared to the single pixel patch attack, namely, only approximately 5\% for ResNet101 (from 94.57\% to 89.85\%, ~\ref{app:table:2}). Finally, the small size of patches with a fixed proportion of damaged pixels allows patches to scatter more across the whole picture, resulting in more uniform perturbation of model filters' receptive fields.

\begin{figure*}[h!]
    \centering
    \begin{subfigure}
        {0.24\textwidth}
        \includegraphics[width=\textwidth]{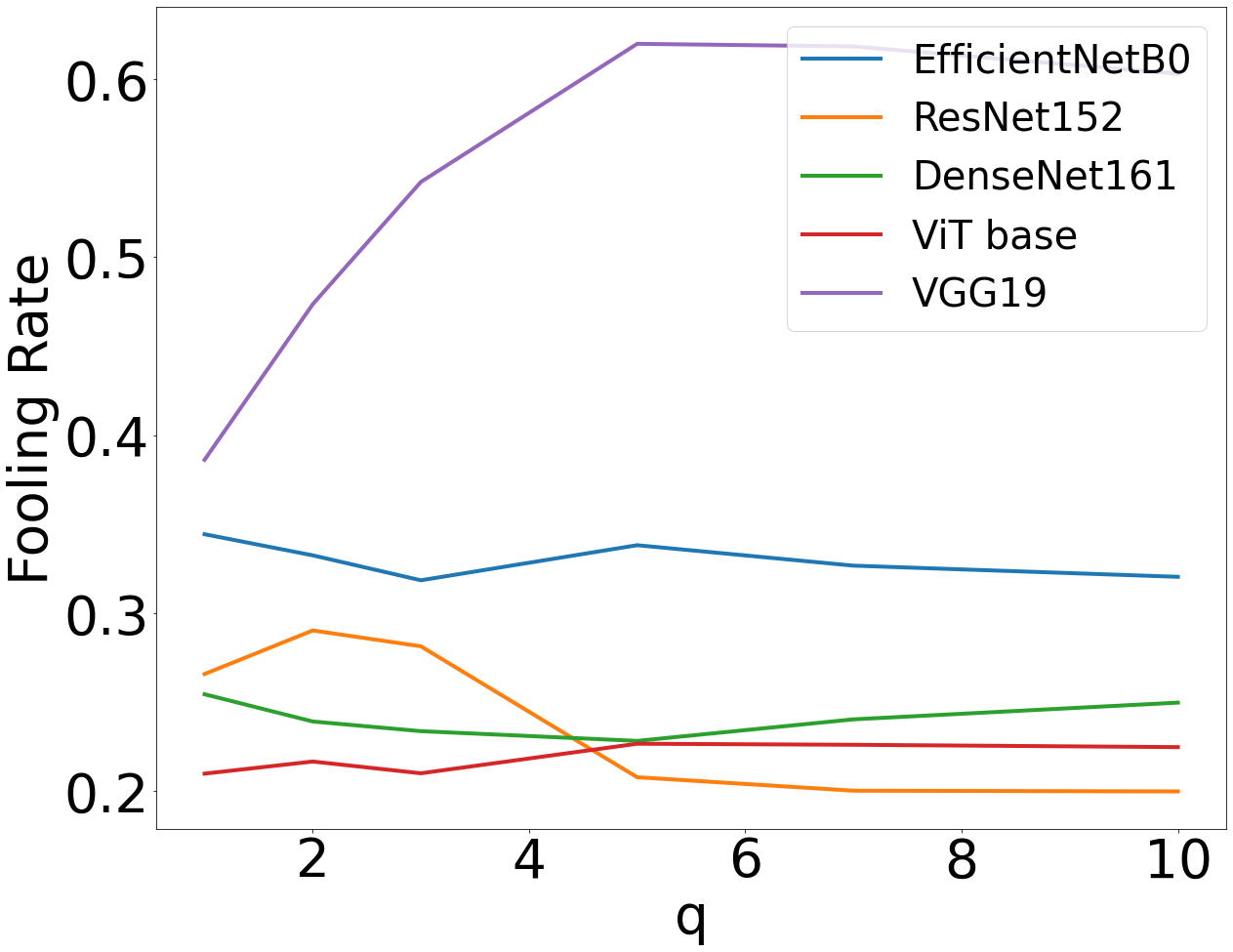}
        \caption{SV Attack.}
    \end{subfigure}
    \begin{subfigure}
        {0.24\textwidth}
        \includegraphics[width=\textwidth]{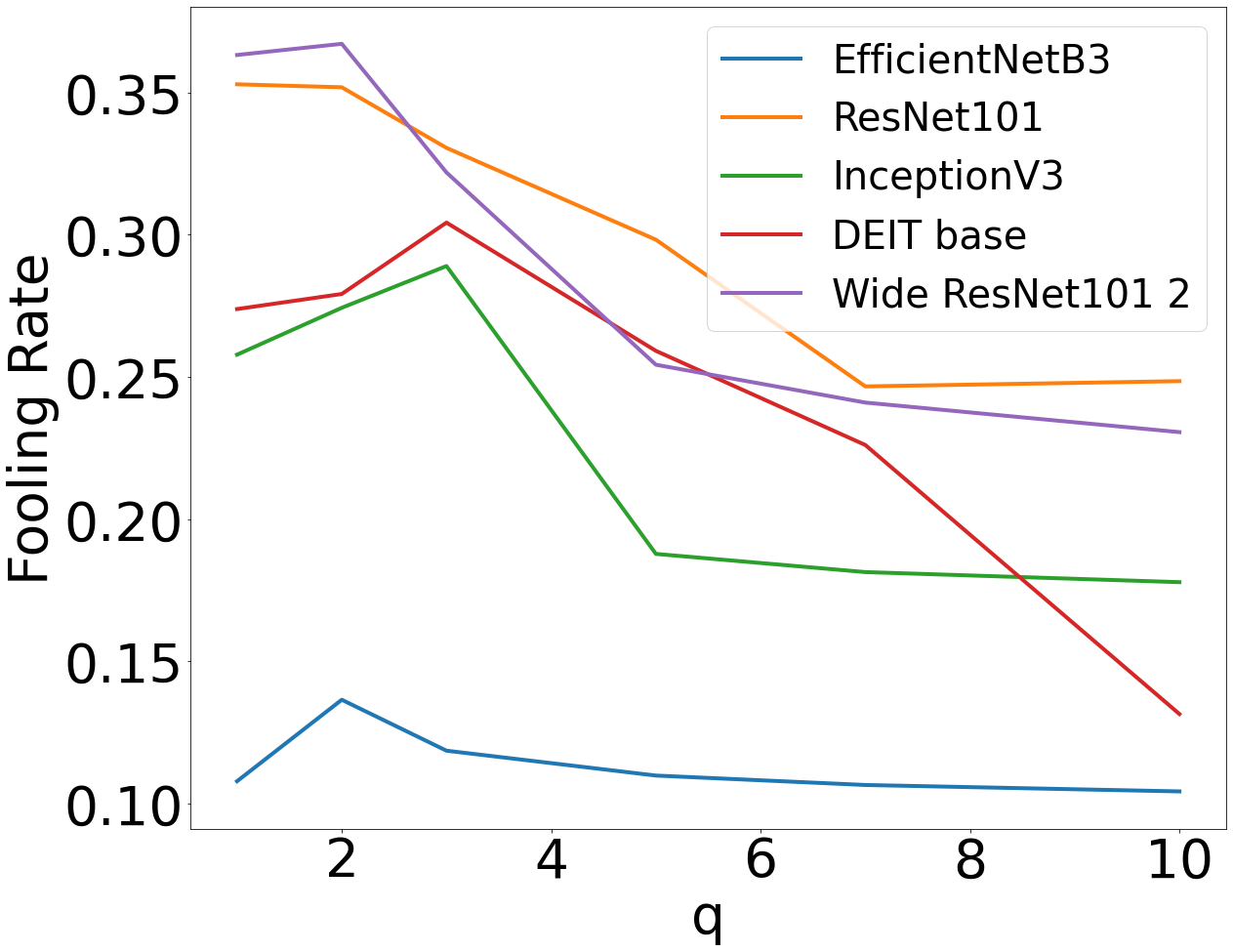}
        \caption{SV Attack.}
    \end{subfigure}
    \begin{subfigure}
        {0.24\textwidth}
        \includegraphics[width=\textwidth]{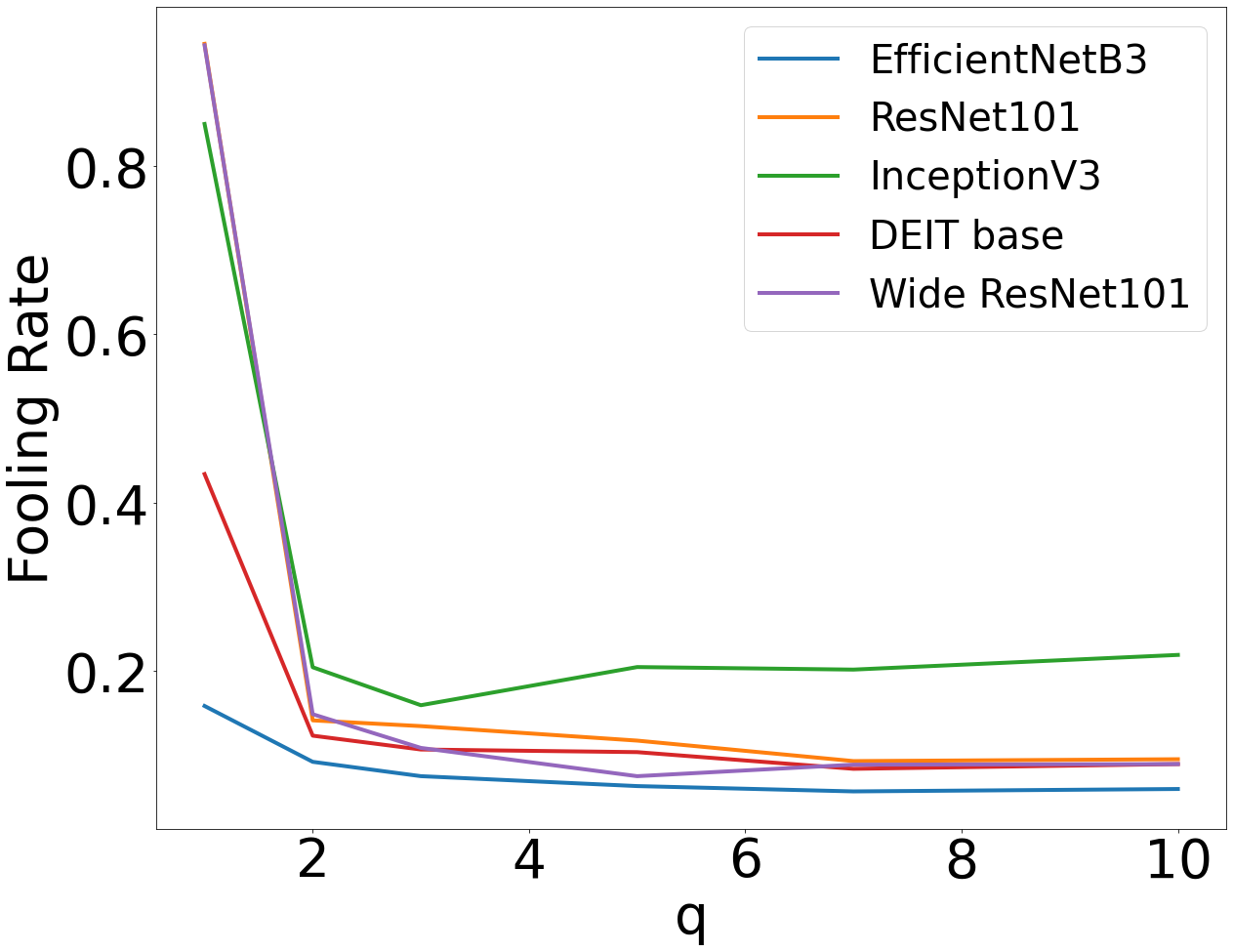}
        \caption{TPower Attack.}
    \end{subfigure}
    \begin{subfigure}
        {0.24\textwidth}
        \includegraphics[width=\textwidth]{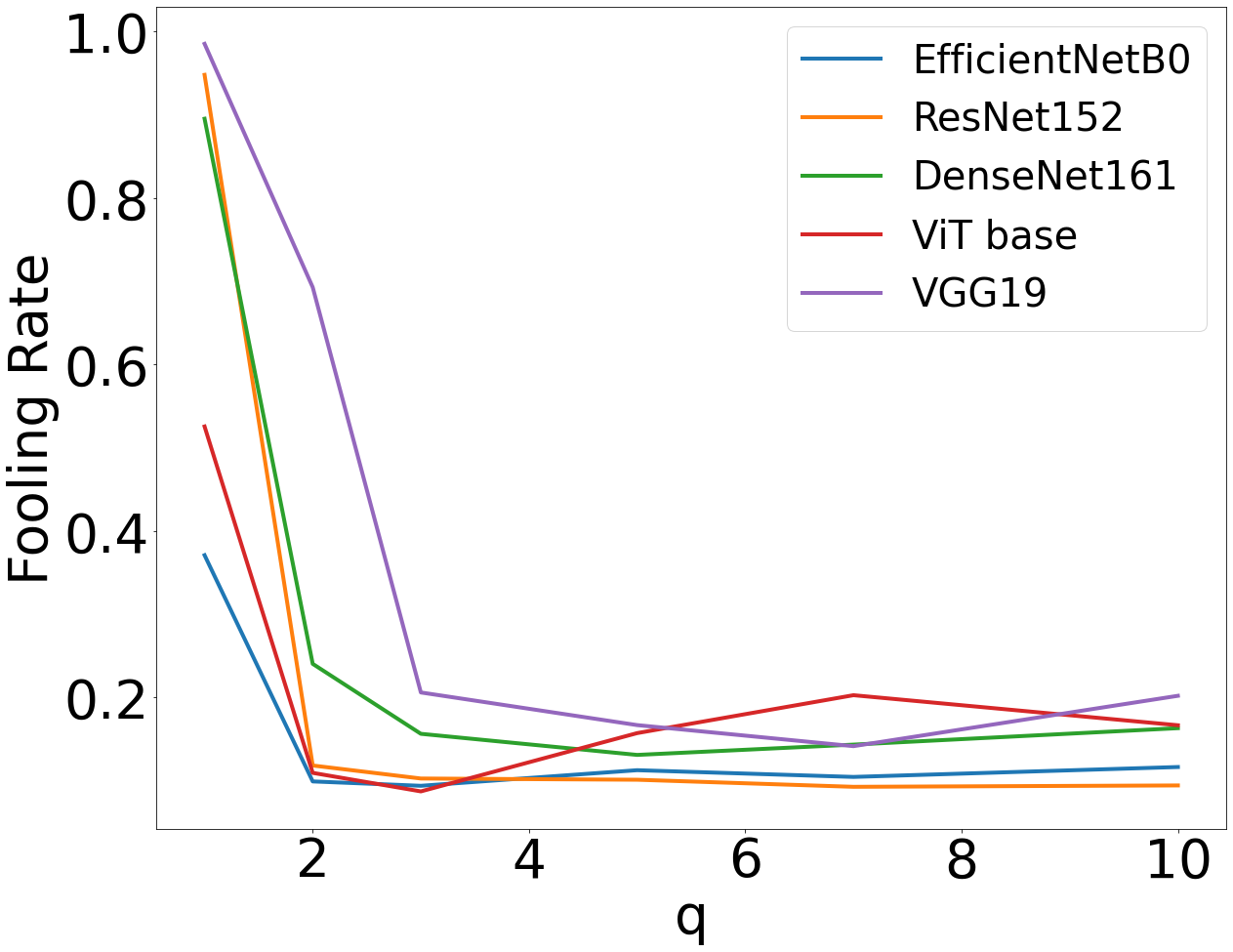}
        \caption{TPower Attack.}
    \end{subfigure}
    \caption{Dependence of Fooling Rate (FR)
    on $q$ for TPower Attack. For sparse attacks, optimal parameters from gridsearch were frozen except for $q$ (see Table \ref{table:1}) and reused for the dense one.
    }
    \label{fig:dependence_fr_q} 
\end{figure*}
\begin{figure*}[h!]
    \centering
    \begin{subfigure}
        {0.16\textwidth}
        \includegraphics[width=\textwidth]{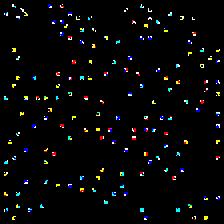}
        \caption{$q = 1$\\TPower Attack}
    \end{subfigure}
    \begin{subfigure}
        {0.16\textwidth}
        \includegraphics[width=\textwidth]{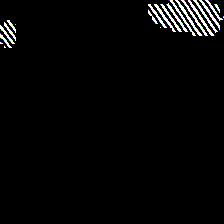}
        \caption{$q = 5$\\TPower Attack}
    \end{subfigure}
    \begin{subfigure}
        {0.16\textwidth}
        \includegraphics[width=\textwidth]{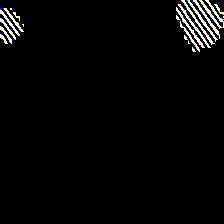}
        \caption{$q = 10$\\TPower Attack}
    \end{subfigure}
    \begin{subfigure}
        {0.16\textwidth}
        \includegraphics[width=\textwidth]{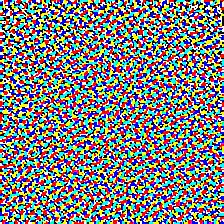}
        \caption{$q = 1$\\SV Attack}
    \end{subfigure}
    \begin{subfigure}
        {0.16\textwidth}
        \includegraphics[width=\textwidth]{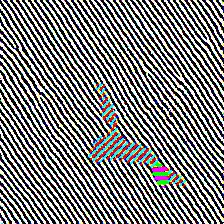}
        \caption{$q = 5$\\SV Attack}
    \end{subfigure}
    \begin{subfigure}
        {0.16\textwidth}
        \includegraphics[width=\textwidth]{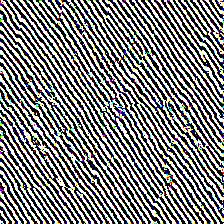}
        \caption{$q = 10$\\SV Attack}
    \end{subfigure}
    \caption{Universal adversarial perturbations constructed for the VGG19 model.}
    \label{fig:vgg}
\end{figure*}

\textbf{Dependence on $q$.}
In~\cite{Khrulkov_2018_CVPR}, on the example of VGG19, authors demonstrate that model vulnerability increases with $q$ and saturates when $q = 5$. 
The last is explained by the fact that $q = 5$ is enough to smooth the approximation of $q = \infty$. 
On the contrary, for the majority of models, we obtained almost opposite results: higher $q$ values are less efficient in terms of FR for both methods, ours and the SV attack (see Figure~\ref{fig:dependence_fr_q}).
However, for the sparse attack setting, all models' dependence becomes unambiguous 
even for the VGG19 model. 
In addition, with $q = 1$, patches are arranged more evenly across the perturbation image than for larger values of $q$, which is depicted in Figure~\ref{fig:vgg}.

\textbf{Dependence on layer number.}
In our experiments, to investigate attack performance depending on a layer, we introduce layer ratio: the layer number normalized to the model depth. From Figure~\ref{fig:layer number}, one can observe that lower layers are more effective, empirically confirming the hypothesis of perturbation propagation through the network and repeating the SV attack property.

\begin{figure}[h!]
    \centering
    \begin{subfigure}
        {0.24\textwidth}
        \includegraphics[width=\textwidth]{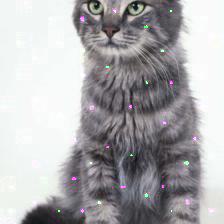}
        \caption{DenseNet161}
    \end{subfigure}
    \begin{subfigure}
        {0.24\textwidth}
        \includegraphics[width=\textwidth]{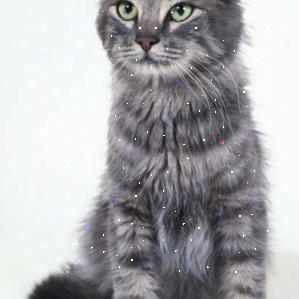}
        \caption{InceptionV3}
    \end{subfigure}
    \begin{subfigure}
        {0.24\textwidth}
        \includegraphics[width=\textwidth]{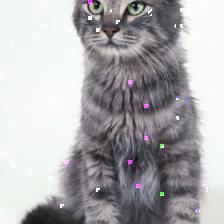}
        \caption{ResNet101}
    \end{subfigure}
    \begin{subfigure}
        {0.24\textwidth}
        \includegraphics[width=\textwidth]{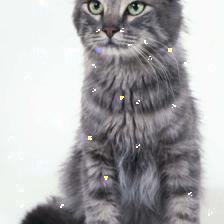}
        \caption{ResNet152} 
    \end{subfigure}
    \caption{
    UAPs and corresponding attacked images obtained using our TPower approach. The $k$ parameter was manually selected such that sparse UAPs reach approximately the same validation Fooling Rate (FR)
    as SV attacks.}
    \label{fig:top-k images}
\end{figure}

\textbf{Cardinality experiments.} 
We analyzed one of the critical hyperparameters - the number of adversarial patches denoted $k$. This hyperparameter plays a pivotal role in determining the ratio of damaged pixels of the attack, as well as the overall performance of the attack. In the initial experiments, we selected the $k$ parameter following the 5\% rate of affected pixels, producing promising results on our dataset. 
We conducted an additional experiment to determine how many sparse adversarial patches are enough to obtain the same fooling rate as for the dense attack. Figure \ref{fig:top-k images} illustrates the resulting images for four models. As anticipated, the choice of $k$ significantly affects the attack performance. However, from Table~\ref{table:pixels_pers}, one can conclude that less than 1\% of pixels is enough to obtain an equally efficient attack with SV one.
\begin{table*}[h!]
    \centering
    \begin{tabular}{lclc}
         \toprule
         Model & \% & Model & \% \\
         \midrule
         InceptionV3 & 0.56 & ResNet152 & 0.27 \\
         DenseNet161 & 0.5 & ResNet101 & 0.37 \\
         \bottomrule
    \end{tabular}
    \caption{The percentage of pixels required for TPower attack to achieve approximately the same FR as SV attack (see Table~\ref{table:2}).}
\label{table:pixels_pers}
\end{table*}

\textbf{Median filtration.} As mentioned above, the constructed perturbations consist of full-magnitude damaged patches scattered uniformly on the image. Due to the small patch size, one can propose median filtration of the vanilla method to mitigate such attack influence. 
Consequently, we have conducted experiments on the median filtration of attacked images with different window sizes. From Table~\ref{tab:filtration}, we observe a decrease in FR, e.g., for EfficientNetB3 and $3\times3$ filter, we get a $1 / 3$ decrease for FR from the initial one; for some models like DenseNet161, the FR decreases to only 79\%. 
However, as a hyperparameter, the filter size should be selected for each model and balance between efficient filtration and over-blurring. 

\begin{table*}[!ht]
    \centering
    \begin{tabular}{l|cccccc}
    \toprule
        Model & 3x3 & 5x5 & 7x7 & 11x11 & 15x15\\ \midrule
        DN & 95.32 & 97.03 & 94.97 & 79.66 & 88.25\\
        ENB0 & 17.31 & 29.27 & 40.99 & 66.95 & 82.34\\
        ENB3 & 9.13 & 16.54 & 24.07 & 41.70 & 59.33 \\ 
        \bottomrule
    \end{tabular}
    \caption{
    Fooling Rate (FR) 
    after the median filtration results for three models: EfficientNetB0 (ENB0), EfficientNetB3 (ENB3) and DenseNet161 (DN). We see that median filtration helps to eliminate attacks, but the optimal window size is not the same for all models and should be tuned. Moreover, exceeding the optimal threshold results in over-blurring and a decrease in the performance of the model, not due to the attack but because of the bad quality of the images themselves.}
    \label{tab:filtration}
\end{table*}

To conclude, the median filter can make the attack harder to fool the victim model but does not protect from it entirely. 
More reliable way to protect models is to use attack detectors or/and robust normalizations inside the models; this requires additional training for each attack type which is impractical.

\begin{table*}[h!]
\centering
\begin{tabular}{@{}l|ccccccccc@{}}
\toprule
\multirow{1}{*}{From/To} & \multicolumn{1}{c}{DN} & \multicolumn{1}{c}{ENB0} & \multicolumn{1}{c}{ENB3} & \multicolumn{1}{c}{IncV3} & \multicolumn{1}{c}{RN101} & \multicolumn{1}{c}{RN152} & \multicolumn{1}{c}{WRN101} & \multicolumn{1}{c}{DEIT} & \multicolumn{1}{c}{VIT} \\
 \midrule
 DenseNet161 (DN) & - & 23.89 & 7.68 & 90.09 & 17.31 & 30.76 & 24.59 & 26.76 & 26.29\\ 
 EfficientNetB0 (ENB0) & 27.72 & - & 9.74 & 24.45 & 28.72 & 25.34 & 27.28 & 19.7 & 17.22 \\
 EfficientNetB3 (ENB3) & 12.24 & 12.46 & - & 9.82 & 14.34 & 14.46 & 12.98 & 10.9 & 8.56 \\
 InceptionV3 (IncV3) & 30.78 & 28.92 & 11.64 & - & 27.28 & 24.88 & 25.63 & 26.8 & 18.95 \\
 ResNet101 (RN101) & 21.08 & 21.02 & 8.93 & 11.45 & - & 26.53 & 24.58 & 19.61 & 10.22 \\
 ResNet152 (RN152) & 28.04 & 35.17 & 9.83 & 88.11 & 23.06 & - & 29.52 & 18.53 & 9.9 \\
 Wide ResNet101 (WRN101) & 24.36 & 23.3 & 7.77 & 15.57 & 32.95 & 28.84 & - & 17.17 & 20.63 \\
 DEIT & 17.82 & 17.24 & 7.31 & 9.65 & 25.57 & 18.88 & 20.33 & - & 9.8 \\
 ViT & 23.87 & 23.13 & 8.5 & 14.28 & 28.29 & 21.06 & 20.96 & 13.64 & - \\
\bottomrule
\end{tabular}
\caption{\label{table:3}
Transferability results for SV attack in terms of the Fooling Rate (FR). Rows refer to the model adversarial perturbation was computed on, while columns~---to the victim one on which the attack was tested.}
\end{table*}

\begin{table*}[h!]
\centering
\begin{tabular}{@{}l|ccccccccc@{}}
\toprule
\multirow{1}{*}{From/To} & \multicolumn{1}{c}{DN} & \multicolumn{1}{c}{ENB0} & \multicolumn{1}{c}{ENB3} & \multicolumn{1}{c}{IncV3} & \multicolumn{1}{c}{RN101} & \multicolumn{1}{c}{RN152} & \multicolumn{1}{c}{WRN101} & \multicolumn{1}{c}{DEIT} & \multicolumn{1}{c}{VIT} \\
 \midrule
 DenseNet161 (DN) & - & 15.29 & 8.88 & 55.31 & 83.9 & 68.93 & 92.98 & 16.66 & 17.29 \\
 EfficientNetB0 (ENB0) & 78.51 & - & 9.41 & 41.58 & 88.44 & 77.42 & 78.67 & 23.03 & 27.78 \\
 EfficientNetB3 (ENB3) & 75.82 & 28.2 & - & 41.02 & 76.74 & 66.03 & 71.45 & 24.61 & 29.81 \\
 InceptionV3 (IncV3) & 93.94 & 35.9 & 14.65 & - & 98.01 & 95.92 & 96.62 & 31.18 & 58.31 \\
 ResNet101 (RN101) & 90.09 & 18.52 & 9.65 & 60.67 & - & 88.08 & 96.48 & 17.42 & 23.12 \\
 ResNet152 (RN152) & 84.39 & 22.05 & 10.21 & 60.41 & 96.98 & - & 94.94 & 21.11 & 29.61 \\
 Wide ResNet101 (WRN101) & 86.55 & 16.68 & 8.64 & 56.43 & 90.7 & 79.34 & - & 17.45 & 21.12 \\
 DEIT & 75.36 & 40.64 & 10.64 & 51.32 & 75.97 & 71.12 & 80.2 & - & 31.2 \\
 ViT & 77.27 & 22.12 & 10.19 & 63.72 & 98.28 & 97.19 & 94.85 & 32.79 & - \\
\bottomrule
\end{tabular}
\caption{\label{table:4}
Transferability results for proposed TPower attack in terms of the Fooling Rate (FR). Rows refer to the model adversarial perturbation was computed on, while columns~---to the victim one on which the attack was tested.}
\end{table*}

\textbf{Transferability experiments.} 
Following the above setup, in the transferability task, for each model we only consider the optimal perturbations in terms of FR obtained during the gridsearch. The rest of the evaluations are done on the test subset. 
In contrast to the direct task setting, when the adversarial perturbation is applied to the same model on which it was obtained, the attack should be adjusted to the input size of the victim model. In particular, we preprocess adversaries either centre-cropping or zero-padding them to fulfil the victim model input size restriction.
Even though this affects the attack transferability performance, from Table~\ref{table:3} and Table~\ref{table:4}, one can observe Tpower attack higher transferability across different models in the majority of cases.

Table~\ref{table:4} demonstrates that
the most transferable attacks in terms of the fooling rate are the ones trained on transformers and EffecientNets. Moreover, EffecientNets are the most robust among examined models. 
In addition, for the rest surrogate models, attack is transferred almost equally well only among these architectures.
This behaviour could be explained by EfficientNet's significant differences from all other models, as this architecture was not developed manually but using the AutoML MNAS framework. 
Nevertheless, attacks trained on EfficientNets have a sufficient transferability fooling rate of at least 24\%, preserving the attacked picture's good quality (see Figure~\ref{fig:images}) and outperforming the results for dense perturbations. 
It is worth mentioning that such attacks perform better in the transferability setting than in the direct one.

To sum up, the transferability of the proposed TPower approach makes it promising for a grey-box setting, where an attack is trained on one model and applied to an unknown one.
However, it should be emphasized that a more detailed study and comparison of various DNN architectures should be done.

%% file: sec/4_related.tex
\section{Related Work}

In recent years, there have been numerous advancements in adversarial attacks. The concept of adversarial attacks in deep neural networks (DNNs) was first introduced in \cite{szegedy2013intriguing}. To exploit this problem, Szegedy et al. used the L-BFGS algorithm. 
This finding significantly impacted the vision research community, which had previously believed that deep visual features closely approximated the perceptual differences between images using euclidean distances.

The Projected Gradient Descent (PGD) attack is regarded as one of the most potent attacks in the field \cite{madry2017towards}. Their main contribution was to examine the adversarial robustness of deep models using robust optimization, which formalizes adversarial training of deep models as a min-max optimization problem.

Widely used adversarial machine learning techniques such as FGSM and DeepFool are designed to generate perturbations that target individual images to attack a specific network. In contrast, the universal adversarial perturbations~\cite{Moosavi-Dezfooli_2017_CVPR} method introduces a unique approach by creating perturbations that have the potential ability to attack any image and any network.

Separately, we would like to highlight the class of attacks obtained using generative models \cite{mopuri2018nag, hayes2018learning, poursaeed2018generative, chen2023content}. Such frameworks are well suited for creating attacks because GANs are able to learn the whole distribution of perturbations, and not just one perturbation as in non-generative approaches~\cite{mopuri2018nag}.

To create a universal adversarial perturbation, an iterative approach is usually employed. Similar to the DeepFool algorithm \cite{moosavi2016deepfool}, the authors gradually shift a single data value towards its nearest hyperplane. In the case of UAP, this process is repeated for all input data points, pushing them towards their respective hyperplanes consecutively. The main problem with this approach is that it is computationally expensive.

Recent and relevant research includes \cite{Khrulkov_2018_CVPR}, where the authors introduced a new algorithm for constructing universal perturbations. The algorithm is based on computing the $(p, q)$-singular values of Jacobian matrices of the network's hidden layers. The authors utilized the well-known Power Method algorithm \citep{boyd1974power} to compute the $(p, q)$-singular vectors. This idea has become popular not only in computer vision; the recent work adopts this algorithm for NLP problems as an example \cite{tsymboi2023layerwise}.

Since then, new gradient-based attacks have appeared, like flexible perturbation sets attacks \cite{wong2019wasserstein, wong2020learning}, as well as attacks with access only to the output scores of the classifier \cite{guo2019subspace, cheng2018query, wang2020spanning, guo2019simple, andriushchenko2020square} and decision-based attacks with access only to predicted labels \cite{chen2020hopskipjumpattack}.

Despite the rapid development of this area, universal attacks are still relatively rare. Many of them are based on generative models \cite{sarkar2017upset, mopuri2018nag, poursaeed2018generative, mao2020gap++}. Recently, a trend in this area has appeared related to the representation of a universal attack as a semantic feature \cite{zhang2020understanding}. Following this work, a bridge between universal and non-universal settings was built \cite{li2022learning}.

%% file: sec/5_limitations.tex
\section{Limitations and future work}
One of the restrictions of the proposed approach is the fixed predefined attack cardinality, and due to the lack of convergence, heuristic reduction to this value should be made. One way to overcome this issue is to replace the truncation operator with adaptive threshold shrinkage obtained via the Alternating Direction Method of Multipliers (ADMM, \cite{boyd2011distributed}), which is planned to be done in future work.

Another weakness is that for sparse attacks to be efficient, we need to use higher magnitudes while keeping the percentage of damaged pixels low. As a result, adversarial perturbation could be considered as an outlier with straightforward defence via weights clipping at each neural network layer. However, clipping ranges for this case must be well-estimated. We need to tune these parameters, e.g., estimate it based on statistics of layers outputs which is infeasible in the grey-box setting when the perturbation is transferred between models.

Finally, it would be interesting to investigate 
the sparse attack transferability in more realistic settings when both model and dataset are unknown, revealing task-independent adversarial perturbations.

%% file: sec/6_conclusions.tex
\section{Conclusion}
This paper presents a new approach for sparse universal adversarial attack generation. Following \cite{Khrulkov_2018_CVPR}, we assume that the perturbation of an intermediate layer propagates further through the network. 
The main outcome of the paper is that by using only an additional truncation operator, we can construct the perturbation that will alter at most 5\% of input image pixels without a decrease in fooling rate compared to the dense algorithm version, but with a significant increase. 
Moreover, our attack is still efficient regarding the sample size used for perturbation training. 
In particular, utilizing 256 samples is enough to achieve at least a 50\% fooling rate for the majority of models, with a maximum of 94\%. We perform a comprehensive study of 10 architectures, revealing their vulnerability to sparse universal attacks. We also show that found attack vectors are highly transferable, revealing an extremely high vulnerability of ResNets. Furthermore, our attack can be well generalized across different networks without a decrease in the fooling rate.

%% file: sec/X_suppl.tex
\clearpage

\section{Implementaion details}
\textbf{Image preprocessing.} 
The preprocessing stage is a crucial step in computer vision tasks that involves cleaning, standardizing, and enhancing the input data to improve model performance. In the context of these experiments, the preprocessing pipeline for the ImageNet ILSVRC2012 dataset will be discussed.

At first, we divided the ImageNet dataset between a training subset with 256 images, a validation subset of 5000 images and the rest for the test subset. The second step in the preprocessing pipeline involves resizing the input images to a fixed resolution. This is necessary to ensure that all images have the same dimensions and to reduce the computational overhead of training the models. The images will be resized to a resolution of crop sizes corresponding to the used models.
After that, our actions for the subsets are different. We compute the attack tensor on the normalized train subset. We apply an attack on validation and test subsets.

Overall, the preprocessing pipeline for the ImageNet ILSVRC2012 dataset for our experiments consists of the train-validation-test split, resizing/cropping, attack applying, clipping to [0, 1] (for validation and test subset) and normalization.

\textbf{Layer selection.} We gradually went through all semantic blocks to study the performance dependence on the layer to be attacked:
\begin{itemize}
    \item DenseNet161: dense layers and transition blocks,
    \item EffecientNetB0 and EffecientNetB3: bottleneck MBConv blocks,
    \item InceptionV3: max poolings and mixed blocks,
    \item ResNet101,~ResNet152~and~WideResNet101:
    
    residual blocks,
    \item Vit and DEIT: encoder layers.
\end{itemize}

\section{Experiments}
\begin{figure*}
    \centering
    \begin{subfigure}
        {0.24\textwidth}
        \includegraphics[width=\textwidth]{sec/images/dense_pics/with_trunc_attack_VGG19_q=1_alpha=0.0392156862745098.jpeg}
        \caption{VGG19, $q = 1$}
    \end{subfigure}
    \begin{subfigure}
        {0.24\textwidth}
        \includegraphics[width=\textwidth]{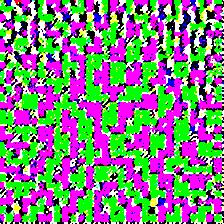}
        \caption{ResNet101, $q = 1$}
    \end{subfigure}
    \begin{subfigure}
        {0.24\textwidth}
        \includegraphics[width=\textwidth]{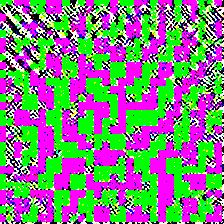}
        \caption{Wide ResNet101, $q = 1$}
    \end{subfigure}
        \begin{subfigure}
        {0.24\textwidth}
        \includegraphics[width=\textwidth]{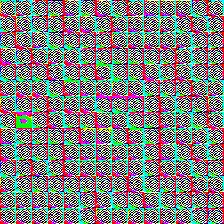}
        \caption{DEIT Base, $q = 1$} 
    \end{subfigure}\\
        \begin{subfigure}
        {0.24\textwidth}
        \includegraphics[width=\textwidth]{sec/images/dense_pics/with_trunc_attack_VGG19_q=5_alpha=0.0392156862745098.jpeg}
        \caption{VGG19, $q = 5$}
    \end{subfigure}
    \begin{subfigure}
        {0.24\textwidth}
        \includegraphics[width=\textwidth]{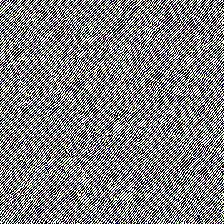}
         \caption{ResNet101, $q = 5$}
    \end{subfigure}
    \begin{subfigure}
        {0.24\textwidth}
        \includegraphics[width=\textwidth]{sec/images/dense_pics/dense_pics_attack_Wide_ResNet101_2.jpeg}  
        \caption{Wide ResNet101, $q = 1$}
    \end{subfigure}
        \begin{subfigure}
        {0.24\textwidth}
        \includegraphics[width=\textwidth]{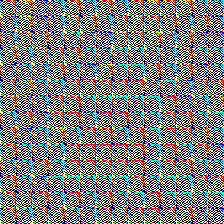}
        \caption{DEIT Base, $q = 3$}
    \end{subfigure}
   \caption{UAPs using SV attack algorithm~\cite{Khrulkov_2018_CVPR}. The first row refers to the fixed parameter value $q = 1$, while the second depicts best-performed perturbations.}
    \label{app:sv_attack_2}
\end{figure*}

\textbf{Grid search results for SV attack.} Optimal hyperparameters with corresponding fooling and attack success rates are presented in Table~\ref{app:table:1}. 
Figure~\ref{app:sv_attack_2} present examples of dense adversarial perturbations. For optimal layers regarding the fooling rate, the dependence on $q$ is ambiguous, while on average, the hypothesis that the greater $q$ is, the better attack performance is\cite{Khrulkov_2018_CVPR} not approved.

\begin{table*}[h!]
\centering
\begin{tabular}{lcccccc}
 \toprule
 Model & $q$ & Attacked Layer & Test ASR & Test FR \\
 \midrule
 DenseNet161 & 3 & features.denseblock2.denselayer10 & 26.49 & 34.25 \\
 EffecientNetB0 & 5 & features.2.1.block & 26.58 & 34.44 \\
 EffecientNetB3 & 2 & features.1.0.block & 8.3 & 13.49 \\
 InceptionV3 & 2 & Mixed\_5b & 19.64 & 27.88 \\
 ResNet101 & 5 & layer1.0 & 43.54 & 50.05 \\
 ResNet152 & 5 & layer1.0 & 28.58 & 35.93 \\
 WideResNet101 & 1 & layer3.1 & 29.24 & 36.35 \\
 DEIT base & 3 & vit.encoder.layer.1 & 23.23 & 31.1 \\
 ViT base & 5 & vit.encoder.layer.0 & 18.75 & 26.01 \\
\bottomrule
\end{tabular}
\caption{Metrics and hyperparameters for the best-performed SV attacks for each model. The attack magnitude was fixed at $\alpha = \frac{10}{255}$.}
\label{app:table:1}
\end{table*}

\textbf{Dependence on patch size.} In general, from Table~\ref{app:table:2}, one can see that pixel-wise attack mode is more efficient regarding the fooling rate. This might be related to the fact that uniform square greed is not an optimal sparsity pattern. However, for most models, the decrease in performance is not dramatic, except for the transformers one.

\begin{table*}[h!]
\centering
\begin{tabular}{lcccc}
 \toprule
 Model & \textbf{Ours} & SV & SGD & LMax \\
 \midrule
 DenseNet161 & \textbf{89.11} & 34.25 & 15.53 & 23 \\
 EffecientNetB0 & \textbf{37.09} & 34.44 & 16.35 & 19.12 \\
 EffecientNetB3 & \textbf{15.22} & 13.49 & 8.22 & 11.21 \\
 InceptionV3 & \textbf{85.04} & 27.88 & 13.11 & 24.64 \\
 ResNet101 & \textbf{94.57} & 50.05 & 17.17 & 46.95 \\
 ResNet152 & \textbf{94.84} & 35.93 & 15.05 & 22.49 \\
 WideResNet101 & \textbf{94.36} & 36.35 & 15.02 & 28.42 \\
 DEIT base & \textbf{43.37} & 31.1 &  19.16 & 23.55 \\
 ViT base & \textbf{52.5} & 26.01 & 17.09 & 28.09 \\
\bottomrule
\end{tabular}
\caption{Metrics for each model with SGD layer maximization attack. The attack magnitude for dense attacks was fixed at $\alpha = \frac{10}{255}$.}
\label{table:lm}
\end{table*}

\begin{figure*}[h!]
    \centering
    \begin{subfigure}
        {0.24\textwidth}
        \includegraphics[width=\textwidth]{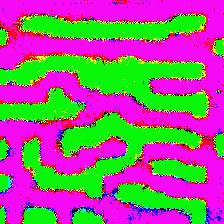}
        \caption{DenseNet161}
    \end{subfigure}
    \begin{subfigure}
        {0.24\textwidth}
        \includegraphics[width=\textwidth]{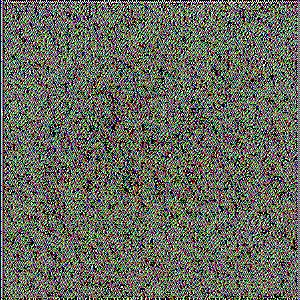}
        \caption{EfficientNetB3}
    \end{subfigure}
    \begin{subfigure}
        {0.24\textwidth}
        \includegraphics[width=\textwidth]{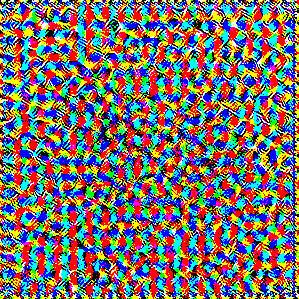}
        \caption{InceptionV3}
    \end{subfigure}
    \begin{subfigure}
        {0.24\textwidth}
        \includegraphics[width=\textwidth]{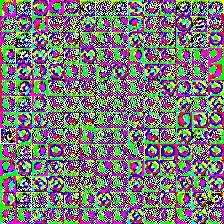}
        \caption{ViT Base}    
    \end{subfigure}\\
        \begin{subfigure}
        {0.24\textwidth}
        \includegraphics[width=\textwidth]{sec/images/appendix/layermax/attack_EfficientNet_B3-2.jpeg}
        \caption{EfficientNetB0}
    \end{subfigure}
    \begin{subfigure}
        {0.24\textwidth}
        \includegraphics[width=\textwidth]{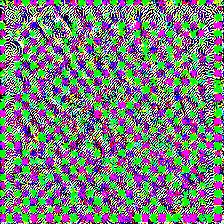}
        \caption{ResNet101}
    \end{subfigure}
    \begin{subfigure}
        {0.24\textwidth}
        \includegraphics[width=\textwidth]{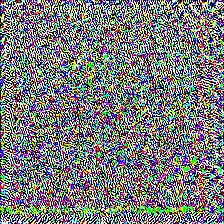}
        \caption{ResNet152}
    \end{subfigure}
    \begin{subfigure}
        {0.24\textwidth}
        \includegraphics[width=\textwidth]{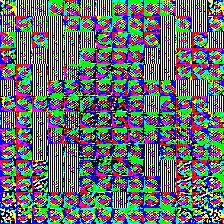}
        \caption{DEIT Base}    
    \end{subfigure}
    \caption{UAPs obtained using SGD with layer maximization attack algorithm~\cite{co2021universal}. Selected layers are the same as optimal in TPower attack.}
    \label{app:sgd_lm_attack}
\end{figure*}

\textbf{SGD with layer maximization.} We also conducted additional experiments and decided to compare with SGD layer maximization attack \cite{co2021universal}, essentially an unlinearized version of our algorithm. The results are presented in \ref{table:lm}, and attack samples are presented in \ref{app:sgd_lm_attack}. As we can observe, layer maximization significantly boosts classic stochastic gradient descent, but the attack still does not reach the performance of our attack or SV attack.

\begin{figure*}[h!]
    \centering
    \begin{subfigure}
        {0.24\textwidth}
        \includegraphics[width=\textwidth]{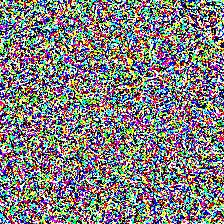}
        \caption{DenseNet161}
    \end{subfigure}
    \begin{subfigure}
        {0.24\textwidth}
        \includegraphics[width=\textwidth]{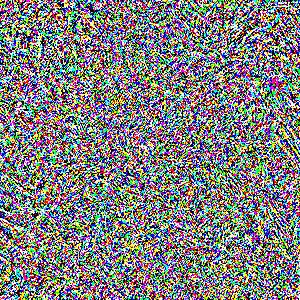}
        \caption{EfficientNetB3}
    \end{subfigure}
    \begin{subfigure}
        {0.24\textwidth}
        \includegraphics[width=\textwidth]{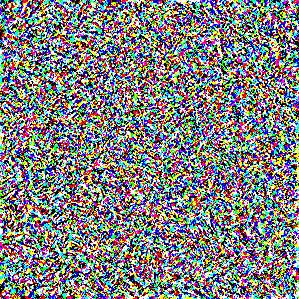}
        \caption{InceptionV3}
    \end{subfigure}
    \begin{subfigure}
        {0.24\textwidth}
        \includegraphics[width=\textwidth]{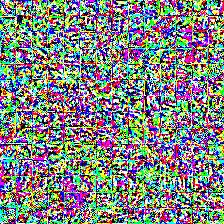}
        \caption{ViT Base}    
    \end{subfigure}\\
        \begin{subfigure}
        {0.24\textwidth}
        \includegraphics[width=\textwidth]{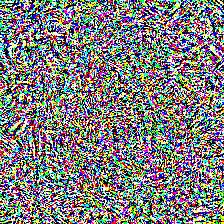}
        \caption{EfficientNetB0}
    \end{subfigure}
    \begin{subfigure}
        {0.24\textwidth}
        \includegraphics[width=\textwidth]{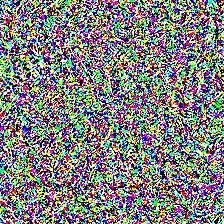}
        \caption{ResNet101}
    \end{subfigure}
    \begin{subfigure}
        {0.24\textwidth}
        \includegraphics[width=\textwidth]{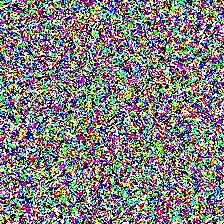}
        \caption{ResNet152}
    \end{subfigure}
    \begin{subfigure}
        {0.24\textwidth}
        \includegraphics[width=\textwidth]{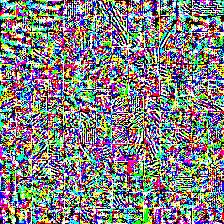}
        \caption{DEIT Base}    
    \end{subfigure}
    \caption{UAPs obtained using SGD attack algorithm~\cite{shafahi2020universal}.}
    \label{app:sgd_attack}
\end{figure*}

\begin{figure*}[h!]
    \centering
    \begin{subfigure}
        {0.24\textwidth}
        \includegraphics[width=\textwidth]{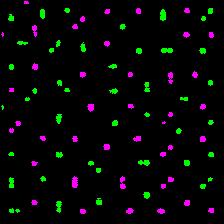}
    \end{subfigure}
    \begin{subfigure}
        {0.24\textwidth}
        \includegraphics[width=\textwidth]{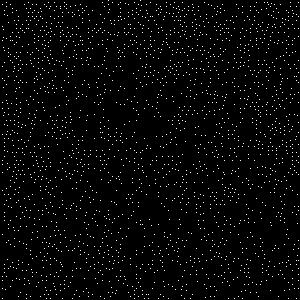}
    \end{subfigure}
    \begin{subfigure}
        {0.24\textwidth}
        \includegraphics[width=\textwidth]{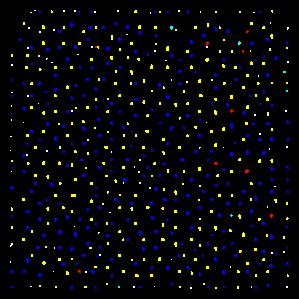}
    \end{subfigure}
    \begin{subfigure}
    {0.24\textwidth}
    \includegraphics[width=\textwidth]{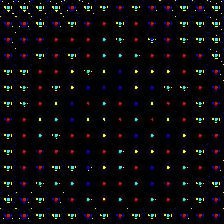}
    \end{subfigure}\\
    \begin{subfigure}
        {0.24\textwidth}
        \includegraphics[width=\textwidth]{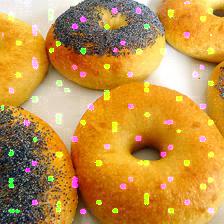}
        \caption{DenseNet161}
    \end{subfigure}
    \begin{subfigure}
        {0.24\textwidth}
        \includegraphics[width=\textwidth]{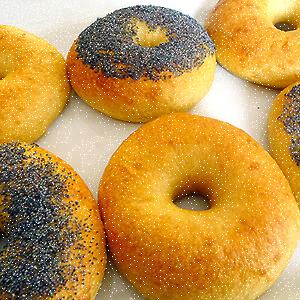}
        \caption{EfficientNetB3}
    \end{subfigure}
    \begin{subfigure}
        {0.24\textwidth}
        \includegraphics[width=\textwidth]{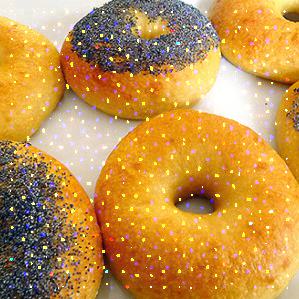}
        \caption{InceptionV3}
    \end{subfigure}
    \begin{subfigure}
    {0.24\textwidth}
    \includegraphics[width=\textwidth]{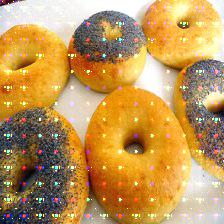}
    \caption{ViT Base} 
    \end{subfigure}\\
    \begin{subfigure}
        {0.24\textwidth}
        \includegraphics[width=\textwidth]{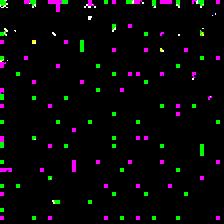}
    \end{subfigure}
    \begin{subfigure}
        {0.24\textwidth}
        \includegraphics[width=\textwidth]{sec/images/truncated_pics/with_trunc_attack_DenseNet161_q=1_alpha=1.jpeg}
    \end{subfigure}
    \begin{subfigure}
        {0.24\textwidth}
        \includegraphics[width=\textwidth]{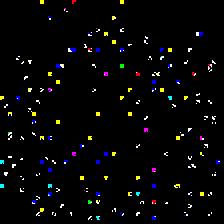}
    \end{subfigure}
    \begin{subfigure}
    {0.24\textwidth}
    \includegraphics[width=\textwidth]{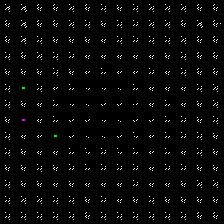}
    \end{subfigure}\\
    \begin{subfigure}
        {0.24\textwidth}
        \includegraphics[width=\textwidth]{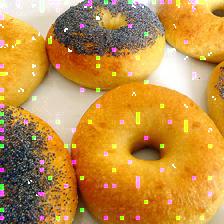}
        \caption{Wide ResNet101}
    \end{subfigure}
    \begin{subfigure}
        {0.24\textwidth}
        \includegraphics[width=\textwidth]{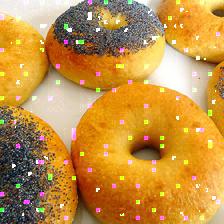}
        \caption{ResNet101}
    \end{subfigure}
    \begin{subfigure}
        {0.24\textwidth}
        \includegraphics[width=\textwidth]{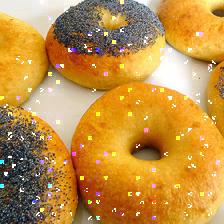}
        \caption{ResNet152}
    \end{subfigure}
    \begin{subfigure}
    {0.24\textwidth}
    \includegraphics[width=\textwidth]{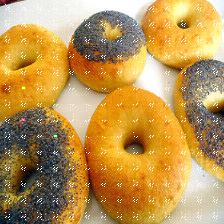}
    \caption{DEIT Base}
    \end{subfigure}
    \caption{UAPs and corresponding attacked images obtained using our TPower approach. Perturbations were computed using the best-performed layers on gridsearch.}
\end{figure*}